\documentclass[sigconf]{acmart}

\usepackage{tikz}
\usepackage{tikzscale}
\usepackage{pgfplots}
\usetikzlibrary{positioning}
\usetikzlibrary{calc}
\setcopyright{none} 
\AtBeginDocument{%
  }

\begin{document}

\title{Fast Converging 3D Gaussian Splatting for 1-Minute Reconstruction}

\author{Ziyu Zhang}
\email{zhangziyu2021@ia.ac.cn}
\affiliation{%
  \institution{University of Chinese Academy of Sciences}
  \institution{Institute of Automation, Chinese Academy of Sciences}
  \city{Beijing}
  \state{Beijing}
  \country{China}
}

\author{Tianle Liu}
\email{tianleliu@whu.edu.cn}
\affiliation{%
  \institution{Wuhan University}
  \city{Wuhan}
  \state{Hubei}
  \country{China}
}
\author{Diantao Tu}
\email{diantao.tu@ia.ac.cn}
\affiliation{%
  \institution{University of Chinese Academy of Sciences}
  \institution{Institute of Automation, Chinese Academy of Sciences}
  \city{Beijing}
  \state{Beijing}
  \country{China}
}
\author{Shuhan Shen}
\email{shshen@nlpr.ia.ac.cn}
\affiliation{%
  \institution{University of Chinese Academy of Sciences}
  \institution{Institute of Automation, Chinese Academy of Sciences}
  \city{Beijing}
  \state{Beijing}
  \country{China}
}

\settopmatter{printacmref=false}  % remove ACM reference format
\renewcommand\footnotetextcopyrightpermission[1]{} % remove copyright footnote
\pagestyle{plain} % remove running headers

\begin{abstract}
  We present a fast 3DGS reconstruction pipeline designed to converge within one minute, developed for \textit{the SIGGRAPH Asia 3DGS Fast Reconstruction Challenge}.
The challenge consists of an initial round using SLAM-generated camera poses (with noisy trajectories) and a final round using COLMAP~\cite{schoenberger2016sfm} poses (highly accurate).
To robustly handle these heterogeneous settings, we develop a two-stage solution.\par
In the \textit{first round}, we use
(1) reverse per-Gaussian parallel optimization and compact forward splatting based on Taming-GS~\cite{10.1145/3680528.3687694} and Speedy-splat~\cite{HansonSpeedy},
(2) load-balanced tiling,
(3) an anchor-based Neural-Gaussian representation~\cite{scaffoldgs} enabling rapid convergence with fewer learnable parameters,
(4) initialization from monocular depth and partially from feed-forward 3DGS models~\cite{jiang2025anysplat}, and
(5) a global pose refinement module for noisy SLAM trajectories.\par
In the \textit{final round}, the accurate COLMAP poses change the optimization landscape; we (1) disable pose refinement, revert from Neural-Gaussians back to standard 3DGS to eliminate MLP inference overhead (2) introduce multi-view consistency-guided Gaussian splitting inspired by Fast-GS~\cite{ren2025fastgs} (3) introduce a depth estimator~\cite{hu2024metric3dv2} to supervise the rendered depth. 
Together, these techniques enable high-fidelity reconstruction under a strict one-minute budget.
Our method achieved the top performance with a PSNR of 28.43 and \textbf{ranked first} in the competition. Code available at \url{https://github.com/will-zzy/siggraph_asia}.
\end{abstract}

\begin{CCSXML}
<ccs2012>
   <concept>
       <concept_id>10010147.10010371.10010352</concept_id>
       <concept_desc>Computing methodologies~Reconstruction</concept_desc>
       <concept_significance>500</concept_significance>
   </concept>
   <concept>
       <concept_id>10010147.10010371.10010396.10010402</concept_id>
       <concept_desc>Computing methodologies~Rendering</concept_desc>
       <concept_significance>300</concept_significance>
   </concept>
   <concept>
       <concept_id>10010147.10010257.10010258.10010262</concept_id>
       <concept_desc>Computing methodologies~Neural networks</concept_desc>
       <concept_significance>300</concept_significance>
   </concept>
   <concept>
       <concept_id>10010147.10010178.10010224</concept_id>
       <concept_desc>Computing methodologies~Computer vision problems</concept_desc>
       <concept_significance>300</concept_significance>
   </concept>
</ccs2012>
\end{CCSXML}

\ccsdesc[500]{Computing methodologies~Reconstruction}
\ccsdesc[300]{Computing methodologies~Rendering}
\ccsdesc[300]{Computing methodologies~Neural networks}
\ccsdesc[300]{Computing methodologies~Computer vision problems}

\keywords{3D Gaussian Splatting, Fast Reconstruction, Neural Rendering, Scene Representation}

\maketitle

\section{Introduction}\par
Efficient reconstruction with 3D Gaussian Splatting~\cite{kerbl3Dgaussians} has become a key research direction.
While existing methods demonstrate impressive visual fidelity, achieving real-time convergence remains challenging, primarily due to (i) the limited efficiency of current rasterization pipelines, (ii) the critical dependence on high-quality initialization, and (iii) the lack of an effective densification and pruning strategy. 
The 3DGS Challenge targets these issues by imposing a strict 1-minute time budget for training.
Our solution is engineered specifically for this constraint, and we describe it in two phases:\par
First Round (SLAM poses):
Noisy poses, sparse point clouds, and challenging frame sampling. Final Round (COLMAP~\cite{schoenberger2016sfm} poses):
Accurate poses, more stable training, and larger allowable Gaussian counts. Below, we summarize the architecture and optimization strategy for each stage.
\section{First Round: Fast Reconstruction with Noisy SLAM Poses}
\subsection{Forward: Compact BBox and Load Balance}\par
GS-like methods~\cite{Yu2024MipSplatting,HansonSpeedy,kerbl3Dgaussians,Huang2DGS2024,zhang2025quadraticgaussiansplattinghigh} involve two main steps during forward rendering. 
The first step, \textit{preprocessCUDA}, can be viewed as analogous to the vertex shader stage in rasterization. It projects each 3D Gaussian onto the image plane using an approximate affine transform, computes the 2D Gaussian centroid and covariance, and then determines the tiles that need to be shaded. Each tile then writes the tile id and the depth of the Gaussian centroid in parallel, which will later be used for depth sorting. \par
\begin{figure}
    \centering
    \includegraphics[width=1\linewidth]{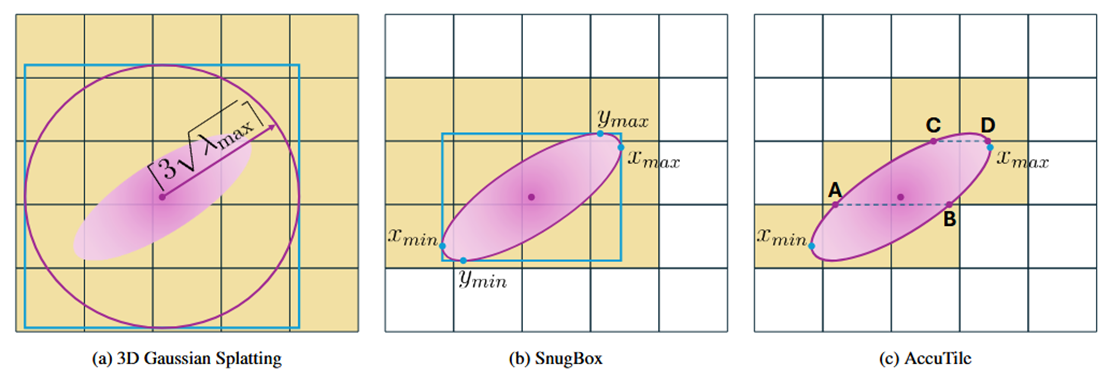}
    \caption{The illustration of the compact bounding strategy, adapted from~\cite{HansonSpeedy}. 
The blue thin boxes denote the bounding boxes, and the yellow regions indicate the tiles being written. 
The first column shows the original 3DGS. 
The second column our method.
The third column presents the pruned bounding box after applying our compact tiling strategy. }
    \label{fig:compact}
    \vspace{-10pt}
\end{figure}

The second step, \textit{renderCUDA}, resembles the fragment shader stage. It processes each pixel in parallel, iterates through the depth-sorted Gaussians, computes the Gaussian weights, performs alpha blending, and outputs the final rendered image.
However, the tile size computed for each Gaussian in \textit{preprocessCUDA} significantly impacts the efficiency of \textit{renderCUDA}. Many of the generated tiles are irrelevant for rendering, resulting in substantial computational waste, especially for elongated Gaussians as shown in Fig.~\ref{fig:compact}. Therefore, we adopt the analytical approach introduced in speedy-splat~\cite{HansonSpeedy} to prune redundant tiles.
Specifically, given the 2D ellipse parameters $Cov2D=\{a,b,c\}$, the ellipse centroid $\mu$, and Gaussian opacity $o$, the ellipse can be parameterized by:
\begin{equation}
2\ln(255\cdot o)=t=ax_d^2+2bx_dy_d+cy_d^2
\end{equation}\par
Using the extremum condition $\partial{y_d}/\partial{x_d}=0$, we can compute the tangent points of the ellipse along the $x$-axis and $y$-axis, i.e. $x_{min}$/$x_{max}$, $y_{min}/y_{max}$. This yields a compact bounding box (the “SnugBox” in Fig.~\ref{fig:compact}):
\begin{equation}
y_{min/max}=\frac{-bx_d\pm\sqrt{(b^2-ac)x_d^2+tc}}{c},\;\;
x_d=\pm\sqrt{\frac{b^2t}{(b^2-ac)a}}
\end{equation}\par
Within the rectangle tiles covered by the SnugBox, we iterate over each column tile and analytically compute its intersection with the ellipse using the left and right tile boundary coordinates (e.g., $x=x_{tmin},x=x_{tmax}$). This allows us to identify which tiles intersect the ellipse and sequentially write out their tile id and depth value.\par
\begin{figure}
    \centering
    \includegraphics[width=1.0\linewidth]{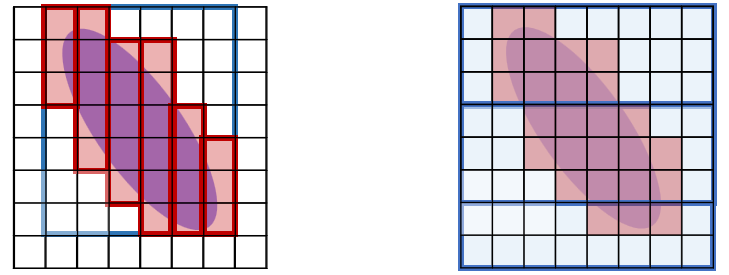}
    \caption{Illustration of sequential writing (left) versus load-balanced writing (right). 
The blue boxes denote the initially recorded tiles, while the red regions indicate the tiles that are finally written.}
    \label{fig:LB_demo}
    \vspace{-10pt}
\end{figure}

Since the BBox sizes vary significantly across Gaussians, the number of tiles processed by each thread also differs greatly, which can lead to thread divergence. Therefore, we also explore using load balance to accelerate this stage. Specifically, we employ cooperative groups to distribute work evenly among threads within the same warp. The difference between this approach and sequential tile writing is illustrated in Fig.~\ref{fig:LB_demo}. In the sequential write mode, a single thread scans through each column or row and analytically computes the intersections between the ellipse and the column/row. In contrast, with load balancing, all threads within the same warp (the blue boxed region) simultaneously evaluate whether their assigned tile intersects with the ellipse.

\subsection{Backward Propagation: Per-Gaussian Parallelism}\par
\begin{figure}[!b]
    \centering
    \begin{minipage}{0.45\linewidth}
        \includegraphics[height=0.5\linewidth]{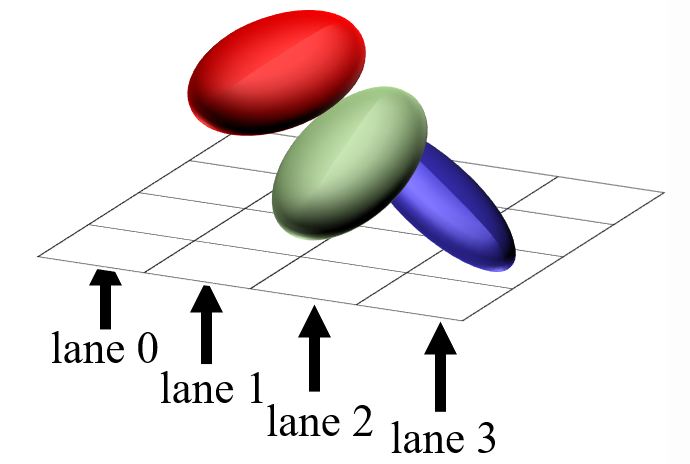}
        \centering
        {\normalsize per-pixel}
    \end{minipage}
    \begin{minipage}{0.45\linewidth}
        \includegraphics[height=0.5\linewidth]{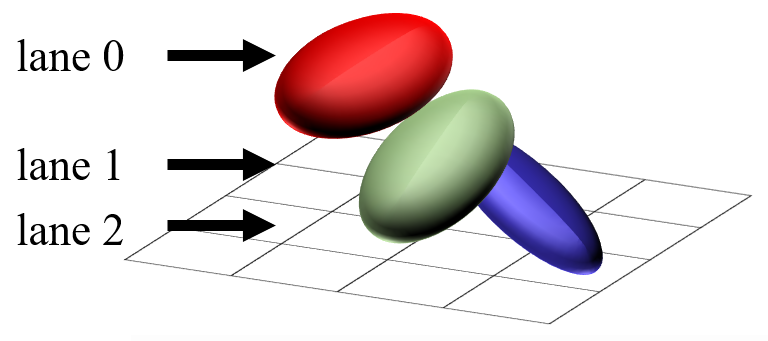}
        \centering
        {\normalsize per-Gaussian}
    \end{minipage}
    \caption{Illustration of per-pixel parallelism (left) and per-Gaussian parallelism (right). 
In per-pixel parallelism, each lane is responsible for processing one pixel, whereas in per-Gaussian parallelism, each lane handles one Gaussian splat.}
    \label{fig:per_gaussian_demo}
\end{figure}

GS-like methods also adopt a two-stage procedure during backpropagation. 
In \textit{preprocessCUDA}, gradients are propagated with a per-Gaussian parallel strategy, while \textit{renderCUDA} uses a per-pixel parallel strategy. However, a single tile (16×16 pixels) often contains hundreds or even thousands of Gaussians. 
In such cases, the available parallelism becomes smaller than the serial workload, causing performance degradation. 
Moreover, during per-pixel backpropagation, multiple pixels may simultaneously write gradients into the same memory region, leading to thread contention and reduced efficiency. 
To address these issues, we adopt the per-Gaussian backward pass introduced in Taming-GS~\cite{10.1145/3680528.3687694}.
Specifically, during the forward pass we record, every 32 splats, the accumulated transmittance $T$, the blended color $C$, and the blended depth $D$. In the backward pass, each warp can then independently perform recursive gradient updates for the splats within its group. The comparison between this strategy and the original per-pixel backward pass is shown in Fig.~\ref{fig:per_gaussian_demo}. Using the cached $T$ and $C$, each warp iterates over all pixels within its tile and recursively accumulates the gradient for each splat. Only after all pixels have been processed does it issue a single atomic write to global memory, significantly reducing thread contention.
Quantitative comparisons between these acceleration strategies are presented in the experimental section.
\subsection{Representation: Neural-Gaussians}\par
\begin{figure}
    \centering
    \begin{tikzpicture}
        \node[inner sep=0pt] (img) {\includegraphics[width=0.9\linewidth]{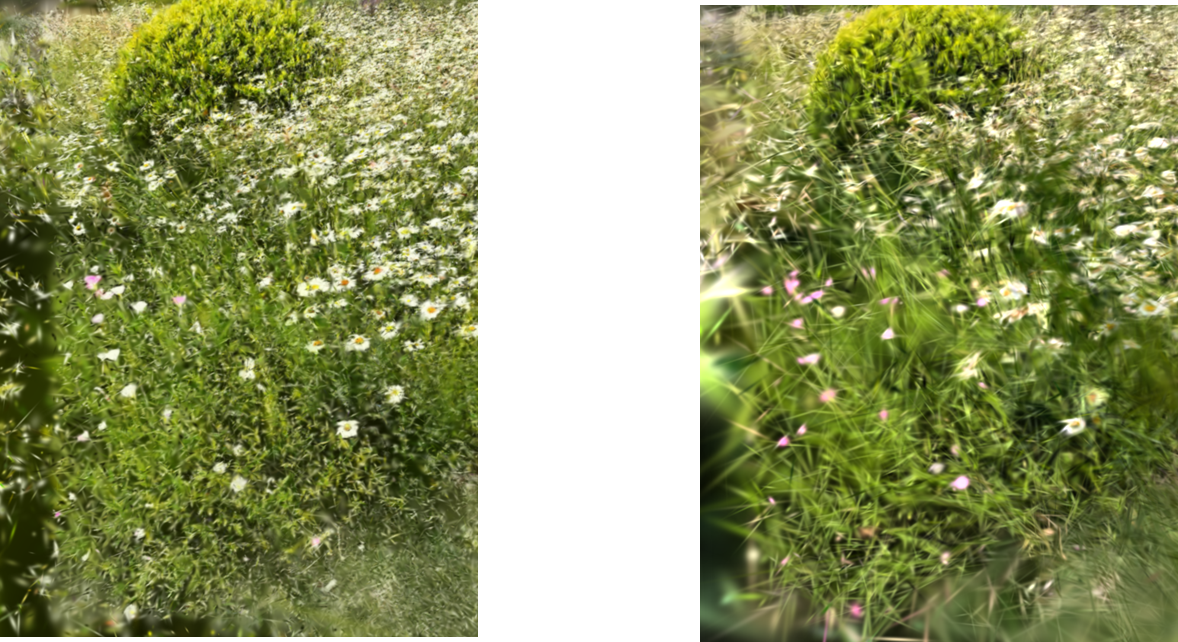}};
        \node[below=2.35cm of img.west, xshift=1.6cm, anchor=center] {\small w/ Neural-Gaussians};
        \node[below=2.35cm of img.west, xshift=6.1cm, anchor=center] {\small w/o Neural-Gaussians};
    
        % RIGHT LABEL – adjust xshift/yshift to fine-tune
        % \node at ($(img.west) + (2.5cm, -0.4cm)$) {\small Left view description};
    \end{tikzpicture}
    
    \caption{Comparison between using Scaffold-GS as the scene representation (left) and the original 3DGS formulation (right). 
The Neural-Gaussian representation leads to faster convergence and captures finer scene details.}
    \label{fig:scaffold}
\end{figure}

After applying the aforementioned acceleration strategies, we observed that the reconstruction still could not converge within one minute on an RTX 4090. Even with an early stopping schedule (20,000 iterations), the runtime remained above the one-minute limit. This motivated us to explore alternative scene representations that enable faster convergence, as shown in Fig.~\ref{fig:scaffold}.\par
Our analysis indicates that in the original formulation, each splat is treated as an independent leaf node, preventing splats from sharing optimization signals. This leads to inefficient learning. Moreover, scenes containing millions of Gaussian primitives introduce a massive number of parameters, further hindering rapid convergence. To address these limitations, we adopt the Neural-Gaussians representation proposed in Scaffold-GS~\cite{scaffoldgs} , where each splat is inferred from an anchor feature rather than optimized directly. This significantly reduces the number of learnable parameters and enables faster convergence.\par
Concretely, we initialize the sparse point cloud as anchors equipped with learnable features and maintain a shallow global MLP. During each forward pass, we feed the anchor features into the MLP to infer the attributes of their associated child Gaussians. These Gaussians are then rendered in the standard manner, and gradients are propagated back to update only the anchor features and the shared MLP. Quantitative comparisons between these initialization strategies are presented in the experimental section.
\subsection{Feedforward Initial Points}
\begin{figure}
    \centering
    \begin{tikzpicture}
        \node[inner sep=0pt] (img) {\includegraphics[width=0.9\linewidth]{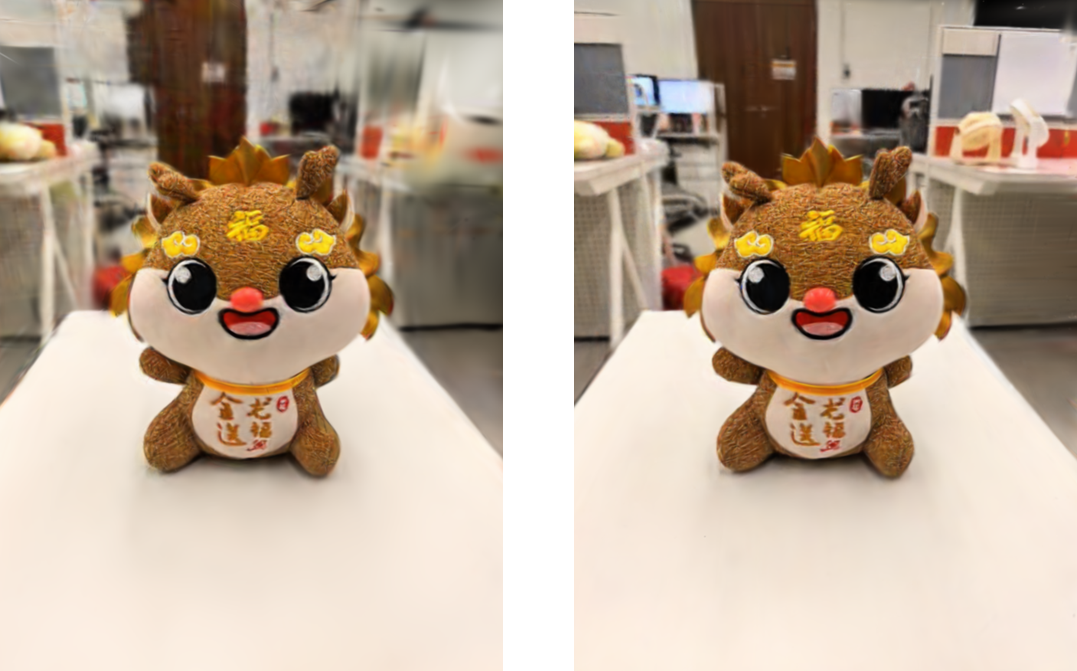}};
        \node[below=2.65cm of img.west, xshift=1.9cm, anchor=center] {\small w/o initialization};
        \node[below=2.65cm of img.west, xshift=5.8cm, anchor=center] {\small w/ initialization};
    
        % RIGHT LABEL – adjust xshift/yshift to fine-tune
        % \node at ($(img.west) + (2.5cm, -0.4cm)$) {\small Left view description};
    \end{tikzpicture}
    \caption{Comparison between non-densified initial point clouds (left) and densified initial point clouds (right). 
Densifying the initialization introduces more geometric details and improves scene coverage.}
    \label{fig:densify}
\end{figure}

We observed that the sparse point clouds provided in the first round dataset were insufficient, leading to slow convergence. To address this issue, we explored two strategies for improving anchor initialization:
(1) We applied Metric3D-v2~\cite{hu2024metric3dv2} to estimate monocular depth, aligned the predicted depth scale to the SLAM point cloud using RANSAC, and randomly sampled the back-projected 3D points as initial anchor positions.
(2) We randomly sampled Gaussian points inferred by the feedforward 3DGS model AnySplat~\cite{jiang2025anysplat} and used them as anchor initialization. 
Fig.~\ref{fig:densify} shows that appropriately increasing the density of the initial point cloud leads to faster convergence. 
Quantitative comparisons between these initialization strategies are presented in the experimental section. 
\subsection{Pose Optimization}
\begin{figure}
    \centering
    \begin{tikzpicture}
        \node[inner sep=0pt] (img) {\includegraphics[width=0.9\linewidth]{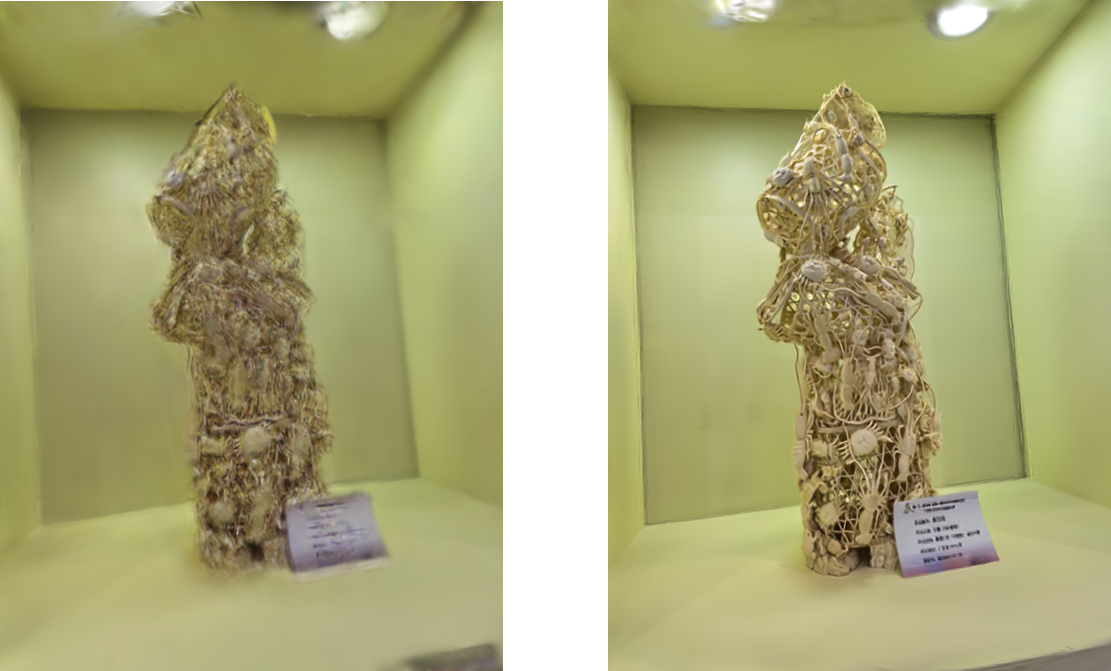}};
        \node[below=2.6cm of img.west, xshift=1.7cm, anchor=center] {\small w/o pose optimization};
        \node[below=2.6cm of img.west, xshift=6.0cm, anchor=center] {\small w/ pose optimization};
    
        % RIGHT LABEL – adjust xshift/yshift to fine-tune
        % \node at ($(img.west) + (2.5cm, -0.4cm)$) {\small Left view description};
    \end{tikzpicture}
    \caption{Comparison between results without pose optimization (left) and with pose optimization (right). 
Applying pose optimization yields visibly sharper and more accurate renderings.}
    \label{fig:pose_opt}
\end{figure}
We observed that the camera poses in the first-round dataset were highly inaccurate, which motivated us to optimize the poses during training. Specifically, we maintain a learnable transform delta $\{\Delta_R, \Delta_t\}$, representing a corrective rotation and translation applied to all cameras. During training, we accumulate the gradient of the rendering loss with respect to this transform delta at every iteration, and update all camera poses using the optimized correction every 300 iterations. 
For simplicity and ease of testing, we maintain a single global transform delta rather than an individual correction for each camera. The same global delta estimated during training is then applied to the test cameras as well. As shown in Fig.~\ref{fig:pose_opt}, pose optimization significantly improves the rendering quality.

\section{Final Round: Fast Reconstruction with Accurate SfM Poses}
The final round dataset contains monocular RGB images together with COLMAP~\cite{schoenberger2016sfm} poses. 
We observed that with these much more accurate camera poses, enabling pose optimization actually degraded the rendering quality, as shown in Table~\ref{tab:final_round}. 
Therefore, the pose refinement module used in the first round was disabled in the final round. 
Meanwhile, we found that integrating anySplat provides high-quality Gaussian initialization, whose attributes are already close to a good solution. 
This reduces the necessity of Neural-Gaussians. 
Moreover, the MLP inference overhead in Scaffold-GS introduces a substantial runtime bottleneck, significantly slowing down training. 
For these reasons, we reverted to the original 3DGS ellipsoidal representation in the final stage. 
Additionally, we introduced several new modules tailored for the final round. We describe these components in detail in the following sections.

\subsection{Depth Regularization}
We supervise the rendered depth using the monocular depth predicted by Metric3D-v2~\cite{hu2024metric3dv2}, which is aligned to the COLMAP scale. 
At the beginning of training, we assign this depth loss a weight of 0.1 to guide the reconstruction toward the correct scene surface more quickly. 
For efficiency reasons, our blended depth corresponds to the centroid depth of each Gaussian rather than the ray–primitive intersection depth, which makes the rendered depth less accurate. 
Therefore, we gradually reduce the weight of this depth loss to zero as training progresses.

Additionally, we supervise disparity (the inverse depth) instead of raw depth values, which helps mitigate the ambiguity and instability of depth estimation in distant regions.

\subsection{Multi-view Score Guided Densification}
To compensate for the reduced convergence speed after removing the Neural-Gaussian representation, we adopt the multi-view score-guided densification and pruning strategy introduced in Fast-GS~\cite{ren2025fastgs}. This strategy provides an efficient mechanism for guiding Gaussian primitives toward faster and more stable convergence.

Specifically, during each densify-and-prune stage, we randomly sample $K=10$ training views and render their corresponding RGB images. 
For densification, we first compute the absolute photometric error between the $j$-th rendered image and the ground-truth image, followed by normalization:
\begin{equation}
    e_{u,v}^j=\mathrm{normalize}(||R^j(u,v)-G^j(u,v)||_1)
\end{equation}
We then identify pixels whose photometric error exceeds a threshold, forming a binary mask $\mathcal{M}_{\mathrm{mask}}^j(u,v)=\mathbb{I}(e^j_{u,v}>\tau)$, and collect all Gaussian primitives contributing to these masked pixels, denoted as $G$. The multi-view consistency score for each primitive $i\in G$ is then computed as:
\begin{equation}
    s_i^+=\frac{1}{K}\sum_{j=1}^K\sum_{u,v}{\mathcal{M}_{\mathrm{mask}}^j(u,v)}
\end{equation}
If $s_i^+$ exceeds a predefined threshold, we clone or split the $i$-th Gaussian primitive to increase representational capacity in regions of high error. 
For pruning, we compute the photometric consistency loss for each view:
\begin{equation}
    E_{\mathrm{photo}}^j=(1-\lambda)L_1^j+\lambda(1-L_{\mathrm{SSIM}}^j)
\end{equation}
and use the error mask to compute a pruning score:
\begin{equation}
    s_i^-=\mathrm{normalize}(\sum_{j=1}^K\sum_{u,v}{\mathcal{M}_{\mathrm{mask}}^j(u,v)}\cdot E_{\mathrm{photo}}^j)
\end{equation}
If $s_i^-$ exceeds its threshold, the i-th Gaussian primitive is pruned.
Together, these multi-view guided densification and pruning rules allow the model to rapidly allocate representational capacity to high-error regions while eliminating redundant primitives, significantly accelerating convergence in the final round.

\section{Experiments}
We conduct all experiments on an RTX 4090 GPU. The maximum number of training iterations is 6,000 (with Neural-Gaussians) for round 1 and 15,000 for round 2 (without Neural-Gaussians). 
If the model fails to converge within the 1-minute time limit, we immediately terminate training, save the current Gaussian point cloud, and record the elapsed training time. \par
For the first round, Table~\ref{tab:speed} compares the training time (30k iterations on the TNT dataset) under four configurations: 
(1) baseline 3DGS, 
(2) per-Gaussian parallel backpropagation (w/ B),
(3) per-Gaussian backpropagation combined with sequential writing of compact bounding boxes (w/ B \& F \& Sq), and
(4) per-Gaussian backpropagation combined with load-balanced writing of compact bounding boxes (w/ B \& F \& LB). \par

Table~\ref{tab:first_round} reports the ablation study of all components used in the \textbf{first round} with \textbf{6,000} training iterations.
We observe that removing pose optimization (w/o Pose Opt) causes a significant drop in rendering quality.
Removing Neural-Gaussians (w/o NG) enables reconstruction within 30 seconds but leads to incomplete convergence, resulting in degraded image quality.
Without densifying the initial point cloud (w/o densify), background regions cannot be effectively represented, as foreground Gaussians fail to split sufficiently into those areas.
With all components enabled (Full), the reconstruction achieves the highest rendering quality. \par
Table~\ref{tab:final_round} reports the ablation study conducted in the \textbf{final round} with \textbf{15,000} training iterations. 
When the initial camera poses are already sufficiently accurate, enabling pose optimization instead degrades rendering quality (w/ Pose Opt). 
Removing monocular depth supervision (w/o Depth Prior) slows down early-stage convergence, which negatively impacts the final rendering quality. 
Using Neural-Gaussians without the multi-view score–guided strategy (w/ NG) often fails to complete the full 15,000 iterations, as training is terminated at the 60-second limit, resulting in inferior reconstruction quality. 
Finally, removing Neural-Gaussians while enabling multi-view score guidance, disabling pose optimization, and incorporating monocular depth supervision (Full) achieves the best overall performance.
\begin{table}[]
    \centering
\resizebox{\columnwidth}{!}{%
    \begin{tabular}{|c|c|c|c|c|c|}
\toprule
TNT(/s $\downarrow$) & Barn & Truck & Ignatius & Meeting & Caterp\\
\bottomrule
    3DGS (baseline)     & 638&611&618&574&615 \\
    w/ B     & 191&173&181&145&183 \\
    w/ B \& F \& Sq& 180&\textbf{159}&\textbf{173}&\textbf{137}&\textbf{171}\\
    w/ B \& F \& LB&\textbf{176}&163&177&141&\textbf{171} \\
    
\bottomrule
    \end{tabular}
}
    \caption{Comparison of training time on the TNT dataset~\cite{knapitsch2017tanks} over 30,000 iterations using three configurations: 
per-Gaussian parallel backpropagation (w/ B), sequential writing of compact bounding boxes (w/ B \& F \& Sq), and load-balanced writing of compact bounding boxes (w/ B \& F \& LB).}
    \label{tab:speed}
    \vspace{-10pt}
\end{table}

\begin{table}[]
    \centering
\resizebox{\columnwidth}{!}{
    \begin{tabular}{|c|c|c|c|c|}
\toprule
 & w/o Pose Opt & w/o densify & w/o NG & Full\\
\midrule
    PSNR $\uparrow$ & 21.15&24.89&23.57&\textbf{25.48} \\
    Time(s)$\downarrow$&58.5&51.3 &\textbf{31.7}&60.0 \\
    
\bottomrule
    \end{tabular}
}
    \caption{Ablation study of the used components on the first round challenge dataset.}
    \label{tab:first_round}
    \vspace{-10pt}
\end{table}

\begin{table}[]
    \centering
\resizebox{\columnwidth}{!}{%
    \begin{tabular}{|c|c|c|c|c|}
\toprule
 & w/ NG & w/o Depth Prior & w/ Pose Opt & Full\\
\bottomrule
    PSNR $\uparrow$ & 26.25&28.61&28.37&\textbf{28.72} \\
    Time(s) $\downarrow$&60.0&\textbf{52.4}&56.3&56.2\\
\bottomrule
    \end{tabular}
}
    \caption{Ablation study of the used components on the final round challenge dataset.}
    \label{tab:final_round}
    \vspace{-10pt}
\end{table}

\bibliographystyle{ACM-Reference-Format}
\bibliography{sample-base}

@String{Computing = "Computing" }

@String{Computer = "{IEEE} Computer" }

@InProceedings{HansonSpeedy,
    author    = {Hanson, Alex and Tu, Allen and Lin, Geng and Singla, Vasu and Zwicker, Matthias and Goldstein, Tom},
    title     = {Speedy-Splat: Fast 3D Gaussian Splatting with Sparse Pixels and Sparse Primitives},
    booktitle = {Proceedings of the Computer Vision and Pattern Recognition Conference (CVPR)},
    month     = {June},
    year      = {2025},
    pages     = {21537-21546},
    url       = {https://speedysplat.github.io/}
}

@inproceedings{10.1145/3680528.3687694,
    author = {Mallick, Saswat Subhajyoti and Goel, Rahul and Kerbl, Bernhard and Steinberger, Markus and Carrasco, Francisco Vicente and De La Torre, Fernando},
    title = {Taming 3DGS: High-Quality Radiance Fields with Limited Resources},
    year = {2024},
    isbn = {9798400711312},
    publisher = {Association for Computing Machinery},
    address = {New York, NY, USA},
    url = {https://doi.org/10.1145/3680528.3687694},
    doi = {10.1145/3680528.3687694},
    booktitle = {SIGGRAPH Asia 2024 Conference Papers},
    articleno = {2},
    numpages = {11},
    series = {SA '24}
}

@article{jiang2025anysplat,
  title={AnySplat: Feed-forward 3D Gaussian Splatting from Unconstrained Views},
  author={Jiang, Lihan and Mao, Yucheng and Xu, Linning and Lu, Tao and Ren, Kerui and Jin, Yichen and Xu, Xudong and Yu, Mulin and Pang, Jiangmiao and Zhao, Feng and others},
  journal={arXiv preprint arXiv:2505.23716},
  year={2025}
}

@inproceedings{scaffoldgs,
  title={Scaffold-gs: Structured 3d gaussians for view-adaptive rendering},
  author={Lu, Tao and Yu, Mulin and Xu, Linning and Xiangli, Yuanbo and Wang, Limin and Lin, Dahua and Dai, Bo},
  booktitle={Proceedings of the IEEE/CVF Conference on Computer Vision and Pattern Recognition},
  pages={20654--20664},
  year={2024}
}

@article{hu2024metric3dv2,
  title={Metric3d v2: A versatile monocular geometric foundation model for zero-shot metric depth and surface normal estimation},
  author={Hu, Mu and Yin, Wei and Zhang, Chi and Cai, Zhipeng and Long, Xiaoxiao and Chen, Hao and Wang, Kaixuan and Yu, Gang and Shen, Chunhua and Shen, Shaojie},
  journal={IEEE Transactions on Pattern Analysis and Machine Intelligence},
  year={2024},
  publisher={IEEE}
}

@article{ren2025fastgs,
  title={FastGS: Training 3D Gaussian Splatting in 100 Seconds},
  author={Ren, Shiwei and Wen, Tianci and Fang, Yongchun and Lu, Biao},
  journal={arXiv preprint arXiv:2511.04283},
  year={2025}
}

@Article{kerbl3Dgaussians,
      author       = {Kerbl, Bernhard and Kopanas, Georgios and Leimk{\"u}hler, Thomas and Drettakis, George},
      title        = {3D Gaussian Splatting for Real-Time Radiance Field Rendering},
      journal      = {ACM Transactions on Graphics},
      number       = {4},
      volume       = {42},
      month        = {July},
      year         = {2023},
      url          = {https://repo-sam.inria.fr/fungraph/3d-gaussian-splatting/}
}

@inproceedings{schoenberger2016sfm,
    author={Sch\"{o}nberger, Johannes Lutz and Frahm, Jan-Michael},
    title={Structure-from-Motion Revisited},
    booktitle={Conference on Computer Vision and Pattern Recognition (CVPR)},
    year={2016},
}

@article{knapitsch2017tanks,
  title={Tanks and temples: Benchmarking large-scale scene reconstruction},
  author={Knapitsch, Arno and Park, Jaesik and Zhou, Qian-Yi and Koltun, Vladlen},
  journal={ACM Transactions on Graphics (ToG)},
  volume={36},
  number={4},
  pages={1--13},
  year={2017},
  publisher={ACM New York, NY, USA}
}

@InProceedings{Yu2024MipSplatting,
    author    = {Yu, Zehao and Chen, Anpei and Huang, Binbin and Sattler, Torsten and Geiger, Andreas},
    title     = {Mip-Splatting: Alias-free 3D Gaussian Splatting},
    booktitle = {Proceedings of the IEEE/CVF Conference on Computer Vision and Pattern Recognition (CVPR)},
    month     = {June},
    year      = {2024},
    pages     = {19447-19456}
}

@inproceedings{Huang2DGS2024,
    title={2D Gaussian Splatting for Geometrically Accurate Radiance Fields},
    author={Huang, Binbin and Yu, Zehao and Chen, Anpei and Geiger, Andreas and Gao, Shenghua},
    publisher = {Association for Computing Machinery},
    booktitle = {SIGGRAPH 2024 Conference Papers},
    year      = {2024},
    doi       = {10.1145/3641519.3657428}
}

@inproceedings{zhang2025quadraticgaussiansplattinghigh,
  author    = {Ziyu Zhang and Binbin Huang and Hanqing Jiang and Liyang Zhou and Xiaojun Xiang and Shunhan Shen},
  title     = {{Quadratic Gaussian Splatting: High Quality Surface Reconstruction with Second-order Geometric Primitives}},
  booktitle = {IEEE International Conference on Computer Vision (ICCV)},
  year      = {2025},
}

\appendix

\end{document}